\documentclass{article}

\usepackage{microtype}
\usepackage{graphicx}
\usepackage{subfigure}
\usepackage{booktabs} 
\usepackage{soul}
\usepackage{worrall}
\usepackage{algorithm2e}

\usepackage{xcolor}

\usepackage{hyperref}

\usepackage[accepted]{icml2019}

\icmltitlerunning{Learning to Convolve: A Generalized Weight-Tying Approach}

\begin{document}

\twocolumn[
\icmltitle{Learning to Convolve: A Generalized Weight-Tying Approach}



\icmlsetsymbol{equal}{*}

\begin{icmlauthorlist}
\icmlauthor{Nichita Diaconu}{am,equal}
\icmlauthor{Daniel Worrall}{am,equal}

\end{icmlauthorlist}

\icmlaffiliation{am}{Philips Lab, University of Amsterdam, Netherlands}

\icmlcorrespondingauthor{Daniel Worrall}{d.e.worrall@uva.nl}

\icmlkeywords{Machine Learning, ICML}

\vskip 0.3in
]



\printAffiliationsAndNotice{\icmlEqualContribution} 

\begin{abstract}
Recent work \cite{CohenW16} has shown that generalizations of convolutions, based on group theory, provide powerful inductive biases for learning. In these generalizations, filters are not only translated but can also be rotated, flipped, etc. However, coming up with exact models of how to rotate a $3\times 3$ filter on a square pixel-grid is difficult. In this paper, we learn how to transform filters for use in the group convolution, focussing on roto-translation. For this, we learn a filter basis and all rotated versions of that filter basis. Filters are then encoded by a set of rotation invariant coefficients. To rotate a filter, we switch the basis. We demonstrate we can produce feature maps with low sensitivity to input rotations, while achieving high performance on MNIST and CIFAR-10.
\end{abstract}

\section{Introduction}
\label{sec:introduction}
Convolutional neural networks (CNNs) are now the model of choice for many perceptual recognition tasks, such as image recognition \cite{RussakovskyDSKS15} or semantic segmentation \cite{CordtsORREBFRS16}. Among all the architectural innovations of recent years, a key component has not changed: the convolution. Convolutions constrain a mapping to have translational symmetry \cite{CohenW16}. This has proven to be a significant inductive bias, and has added bonuses of reducing the number of trainable weights via weight-tying. As such, CNNs have improved sample complexity \cite{SokolicGSR17} and thus better generalization compared to their non-symmetric cousins.

Recently, there has been a flurry of works \cite{CohenW16,CohenW16a,WorrallGTB17,WorrallB18,BekkersLVEPD18,WeilerHS18,WeilerGWBC18,EstevesAMD18} on extending the class of symmetries beyond pixelwise translations. Most notably \citet{CohenW16}'s \emph{group convolutions} extend standard translational convolution to the setting where the symmetry is a discrete algebraic group (explained in Section \ref{sec:group_convolution}). In other words, these are convolutions over invertible transformations, so kernels are not only translated but also rotated, flipped, etc. 

One of the key assumptions with \citet{CohenW16} and associated approaches is that the set of transformations forms a group. We cannot pick an arbitrary set of transformations. For instance, in \citet{CohenW16} the authors choose the group of pixelwise translations, $90^\circ$ rotations, and flips, that is the set of all transformations that map the regular square-lattice into itself; and in \citet{HoogeboomPCW18} the authors consider the set of all transformations that map the hexagonal lattice into itself. However, in general the set of $\frac{2\pi}{N}$ rotations for integer $N$ and pixelwise translations does not form a group because of pixelwise discretization, yet in \citet{BekkersLVEPD18} and \citet{WeilerHS18}, the authors use these sets of transformations. Their line of thought is to model roto-translations in the continuous setting and then discretize \emph{post hoc}. To synthesize rotating a filter, they apply bilinear interpolation \cite{BekkersLVEPD18} or a Fourier-based interpolation scheme \cite{WeilerHS18}. Both of these methods suffer from the fact that the \emph{post hoc} discretization is not accounted for by the interpolation schemes employed. 

Our solution is to learn an interpolation scheme to transform the filters, which accounts for this \emph{post hoc} discretization. Our proposal is to learn a space of filters and transformed versions of this space. A filter at one orientation is defined by its coefficients in a chosen basis. It can then be rotated by changing the basis, while keeping the coefficients unchanged. As a result, this is a generalized version of weight-tying. Since the bases are learned, we do not explicitly define a transformation rule on the filters. This avoids some of the overlooked flaws of handcrafted transformations.

Our contributions are:
\begin{itemize}
    \item We present an expanded definition of the group convolution, with a new proof of equivariance.
    \item We pinpoint the spurious non-equivariance properties of existing roto-translational group CNNs down to the non-unitarity of the filter transformations used.
    \item We present a generalized weight-tying method, where the convolution is learned.
\end{itemize}

In Section \ref{sec:background} we cover background material on group convolutions, in Section \ref{sec:method} we explain our method to learn how to rotate filters, and in Section \ref{sec:experiments} we present experiments on MNIST and CIFAR-10 comparing classification accuracy and robustness under rotations of the input.

\section{Background}
\label{sec:background}
Here, we introduce discrete group convolutions. We present them from a novel perspective for freshness and to elucidate the leap from handcrafted transformations to learning them.

\subsection{Standard convolutions}
The standard translational convolution\footnote{Technically this is a correlation, but we stick to the standard deep learning nomenclature.} has the form
\begin{align}
    [f \star_{\mbb{Z}^d} \psi](g) = \sum_{x \in \mbb{Z}^d} f(x) \psi(x - g), \label{eq:standard_convolution}
\end{align}
where $f$ is a signal (image) and $\psi$ is a filter, both defined on domain $\mbb{Z}^d$. It can be interpreted as a collection of inner products of signal $f$ with $g$-translated versions of filter $\psi$. To emphasize (and generalize) this translational behavior, we rewrite the filter shift using the translation operator $\mathcal{L}_g$:
\begin{align}
    [f \star_{\mbb{Z}^d} \psi](g) &= \sum_{x \in \mbb{Z}^d} f(x) \mc{L}_g[\psi](x) \label{eq:standard_convolution_operator} \\
    \mc{L}_{g}[\psi](x) &= \psi(x - g). \label{eq:translation_operator}
\end{align}
In this form, we notice a characteristic property of the convolution. Each of the responses is indexed by the amount of translation $g$. Notice too that $\mc{L}_g$ is indexed by a translation parameter $g$, so we actually have a set of operators $\{\mc{L}_g\}_{g\in G}$, where $G$ is the set of all transformations. Here, the domain of the output response of the convolution is $G = \mbb{Z}^d$. A fundamental property of the convolution is that it is \emph{equivariant} to translations \cite{KondorT18}. This is the behavior that translations of the input $f$ result in translations of the response $[f \star_{\mbb{Z}^d} \psi]$. We write this as
\begin{align}
   \mc{L}_g [f \star_G \psi] = \mc{L}_g[f] \star_G \psi, \label{eq:equivariance}
\end{align}
which highlights the commutativity between convolution and translation. So convolving a filter with a transformed image yields the same responses as first convolving the signal and then applying a transformation to that response. It turns out that the convolution (and reparametrizations of it) is the only linear map that commutes with translation.

\subsection{Group convolution}
\label{sec:group_convolution}
We can extend the standard convolution by swapping the translation operator for another general transformation operator. We shall use the same notation $\mc{L}_g$ to denote this more general transformation operator and thus Equation \ref{eq:standard_convolution_operator} does not change. The basic form of this new convolution can be seen as a collection of inner products between a signal and generically transformed filters, reminiscent of template matching in classical computer vision \cite{Szeliski11}.

\paragraph{Group Actions}
For the generalized convolution to maintain the equivariance property, we have to restrict the class of transformation operators. Following \citet{CohenW16} we demand the following group properties. For all $g,h,k \in G$:
\begin{itemize}
    \item Closure: $\mc{L}_g\mc{L}_h = \mc{L}_{g\circ h}$, where $\circ:G\times G \to G$ denotes composition.
    \item Associativity: $\mc{L}_g(\mc{L}_h\mc{L}_k) = (\mc{L}_g\mc{L}_h)\mc{L}_k$.
    \item Identity: there exists an $e\in G$ such that $\mc{L}_e[f] = f$.
    \item Inverses: for every $\mc{L}_g$ there exists an $\mc{L}_{g^{-1}}$ such that $\mc{L}_g\mc{L}_{g^{-1}} = \mc{L}_{g^{-1}}\mc{L}_g = \mc{L}_e$.
\end{itemize}
If the transformations satisfy these properties then they are called \emph{group actions} and the set $G$ is called a \emph{group}. Typically, we ignore the notation $g\circ h$ and just write $gh$. Note that we have presented the action as a map from functions to functions, but it can also be applied to points. For instance, we can define an action $\mc{R}_R[x] = Rx$, where $R$ is a rotation matrix. We can then define the rotation of a function $f: \mbb{R}^2 \to \mbb{R}$ via the rotation of its domain
\begin{align}
    \mc{R}_R[f](x) = f(\mc{R}_{R}^{-1}[x]) = f(R^{-1}x).
\end{align}
Here we have overloaded the notation $\mc{R}_R$ to act on functions and points. It should be clear from context, which is which. Notice how when we bring the action inside the function to act on the domain, we use the inverse of the action on points. This maintains the relation $\mc{R}_R\mc{R}_S = \mc{R}_{RS}$:
{
\thickmuskip=2mu
\begin{align}
    &\mc{R}_R[\mc{R}_S[f]](x) = \mc{R}_S[f](\mc{R}_{R}^{-1}[x]) = f(\mc{R}_{S}^{-1}[\mc{R}_{R}^{-1}[x]]) \\
    &= f(\mc{R}_{S^{-1} R^{-1}}[x]]) = f(\mc{R}_{(RS)^{-1}}[x]]) = \mc{R}_{RS}[f](x).
\end{align}
}

\paragraph{Group convolutions}
\emph{Group convolutions} $\star_G$ use group actions of the form $\mc{L}_g[f](x) = f(\mc{L}_{g}^{-1}x)$, where:
\begin{align}
    [f \star_{G} \psi](g) = \sum_{x \in X} f(x) \psi(\mc{L}_g^{-1}[x]). \label{eq:group-convolution-old}
\end{align}
They differ from Equation \ref{eq:standard_convolution_operator} in that the domain $X$ of the signal and filter is not necessarily $\mbb{Z}^d$. Recall, the domain of the response is the group $G$. Since we wish to concatenate multiple group convolutions to form a CNN, we typically have $X=G$. In \citet{KondorT18} and \citet{WeilerGWBC18}, the authors call $X$ a \emph{homogeneous space}.

\begin{figure}
    \centering
    \includegraphics[width=\linewidth]{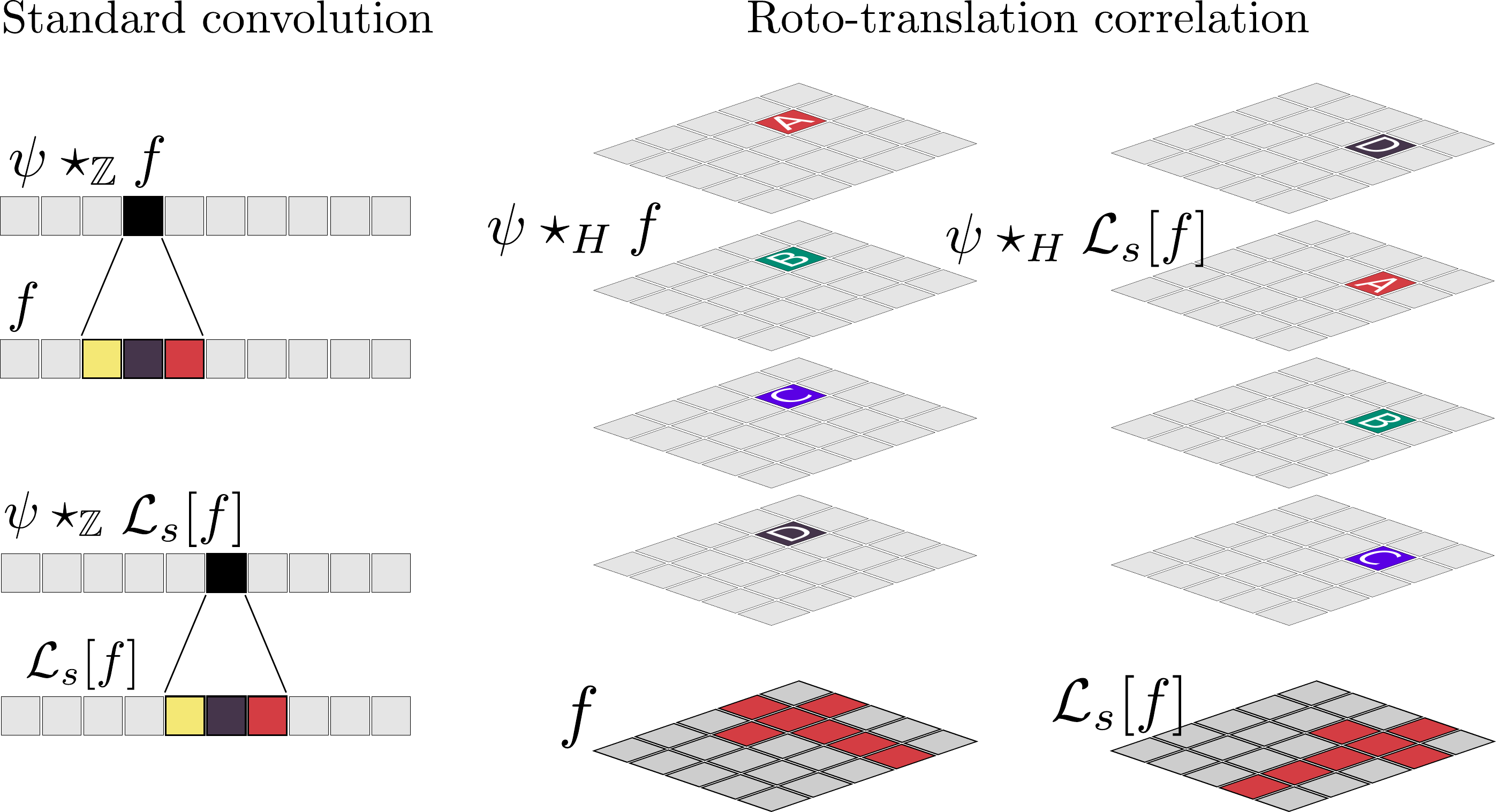}
    \caption{Example of the equivariance properties of a standard convolution versus a roto-translation convolutions. For the standard convolution, pixelwise shifts (left-right translations) of the input induce pixelwise shifts of the response. For the roto-translational group convolution, shifts of the input induce shifts. Rotations of the input induce a rotation of the spatial location of activations within a slice (horizontal plane) at the same time as a rolling (cyclic vertical shift) of the slices. Were we to rotate by ine full turn, the activations would trace a corkscrew path through space.}
    \label{fig:rt-equivariance}
\end{figure}

\paragraph{The roto-translation group}
In this paper, we focus primarily on the roto-translation group; although, our method is applicable to other group-structured transformations. The roto-translation group\footnote{Also known as the Euclidean group} is the group of all proper rigid-body transformations. Transforming a point $x$ is performed as
\begin{align}
    \mc{L}_{R,z}[x] = Rx + z, \label{eq:roto-translation-action}
\end{align}
where $R$ is a rotation matrix and $z$ is the translation vector. We treat the tuple $g=(R,z)$ as a single element in the group $G$. We can also treat the roto-translation action as a composition of two actions: the rotation action $\mc{R}_R[x] = Rx$ followed by the translation action $\mc{T}_z[x] = x + z$, so
\begin{align}
     \mc{L}_{R,z}[x] = \mc{T}_z[\mc{R}_R[x]] = \mc{R}_R[x] + z = Rx + z.
\end{align}
A handy representation of elements $(R,z)$ is the homogeneous representation shown Equation \ref{eq:homogeneous}, where $0$ is an appropriately sized vector of zeros. In this representation, composition of transformations is matrix multiplication. Thus for group elements $(R,z)$ and $(S,x)$ we find the useful expressions
\begin{align}
    \underbrace{\begin{bmatrix}
        R               & z \\
        0^\top     & 1
    \end{bmatrix}}_{(R,z)} \underbrace{\begin{bmatrix}
        S               & x \\
        0^\top     & 1
    \end{bmatrix}}_{(S,x)} &= \underbrace{\begin{bmatrix}
        RS    & Rx+z \\
        0^\top     & 1
    \end{bmatrix}}_{(RS, Rx + z)} \label{eq:homogeneous} \\ \underbrace{\begin{bmatrix}
        R          & z \\
        0^\top     & 1
    \end{bmatrix}^{-1}}_{(R,z)^{-1}} &= \underbrace{\begin{bmatrix}
        R^{-1}          & -R^{-1}z \\
        0^\top     & 1
    \end{bmatrix}}_{(R^{-1}, -R^{-1}z)}.
\end{align}
We can now use this knowledge to write down an explicit group convolution for roto-translation. In both convolutions we will explicitly separate out the filter rotation component, which we later intend to learn. There is one expression for the network input, where $X=\mbb{Z}^d$, and another for intermediate activations, where $X=G$. For the input:
\begin{align}
    [f \star_{G} \psi](R,z) &= \sum_{x \in \mbb{Z}^d} f(x) \psi(R^{-1}(x-z)) \label{eq:input-conv} \\
    &= [f \star_{\mbb{Z}^d} \mc{R}_R[\psi]](z).
\end{align}
In the first line, we have written the group convolution of Equation \ref{eq:group-convolution-old}, substituting in the roto-translation action of Equation \ref{eq:roto-translation-action}. In the second line, we have rewritten this as a standard translational convolution with different rotated versions of the filters $\psi$. For the intermediate convolutions, the behavior is more sophisticated. Since the input activations live on the group, the filters also must live on the group and thus have a rotation and a translation `axis' and are written $\psi(S,x)$. In the following, we shall refer to $\psi_S(\cdot) = \psi(S,\cdot)$ as a \emph{slice} (see Figure \ref{fig:rt-equivariance}):
{
\medmuskip=0mu
\thinmuskip=0.5mu
\thickmuskip=0mu
\begin{align}
    [f \star_{G} \psi](R,z) &= \sum_{(S,x) \in G} f(S,x) \psi(R^{-1}S,R^{-1}(x-z)) \label{eq:intermediate-conv} \\
    &= \sum_{S} [f_S \star_{\mbb{Z}^d} \mc{R}_R[\psi_{R^{-1}S}]](z)
\end{align}
}

Again in the first line we wrote out the convolution using the substitution of Equation \ref{eq:roto-translation-action} in \ref{eq:group-convolution-old}. In the second line we notice that this convolution has two components i) for fixed $S$, we perform a standard translational convolution with feature map slice $f_S$ and an $R$-rotated filter slice $\psi_{R^{-1}S}$, ii) we then sum over all responses from each slice. Notice that on top of the filter rotation we also permute the slices according to $R^{-1}S$, which we call a \emph{roll} (see Figure \ref{fig:rt-equivariance}), since in 2D this corresponds to a cyclic permutation of the slices. It is more complicated in higher dimensions, as shown in \citet{WorrallB18}. 

The equivariance properties of this convolution are interesting. Input shifts induce shifts of the response. Rotations induce a more complex transformation. We visualize this transformation in Figure \ref{fig:rt-equivariance}. First of all, responses are rotated spatially within a slice by the same degree as the input rotation, and at the same time there is a rolling of the slices over the `rotation-axis' of the response tensor. After one full turn the activations return to where they begun, tracing a corkscrew-like path through space.

Our presentation of the group convolution differs from previous works in that we present the group action as $\mc{L}_g[f](x)$ rather than $f(\mc{L}_g^{-1}[x])$. The difference between these two expressions is that in the first, we can transform both the pixel locations and values of a signal, but in the second we can only permute their locations. To demonstrate that this is a more accurate model of signal transformations consider the case where $x$ is a point in the lattice $\mbb{Z}^d$. The transformation $\mc{L}_g^{-1}[x]$ may not actually exist for general $g$---an example is where $g$ represents a $45^\circ$ rotation. Here $\mc{L}_g^{-1}[x]$ maps $x$ to a point off the lattice and therefore cannot exist. To remedy this, we have two options: i) in \citet{CohenW16} the authors opt not to allow transformations where $g^{-1}x$ do not exist and focus on $90^\circ$ rotations, ii) in \citet{BekkersLVEPD18} and \citet{WeilerHS18} the authors instead choose to use an interpolation function to perform filter rotation. This second method is the only satisfactory method to achieve equivariance at non-$90^\circ$ rotations, but this is implementing $\mc{L}_g[f]$, not $f(\mc{L}_g^{-1}[x])$. 

\section{Method}
\label{sec:method}
In the following, we present a modified and expanded definition of the discrete group convolution, based of actions of the form $\mc{L}_g[f]$ rather than actions $f(\mc{L}_g^{-1}[x])$. We then set out the condition needed for this expanded definition to satisfy equivariance, namely that the action has to be unitary under the inner product of the convolution. We then explore what happens if we replace the group for an arbitrary non-group-structured set of transformations.

\subsection{The Unitary Group Convolution}
Our expanded definition of the group convolution uses function transformations of the form $\mc{L}_g[f]$ so that we have
\begin{align}
    [f \star_{G} \psi](g) = \sum_{x \in X} f(x) \mc{L}_g[\psi](x). \label{eq:group-convolution-new}
\end{align}
This may look like a small change from Equation \ref{eq:group-convolution-old}, but it heralds a shift in the nature of the group convolution. It turns out the proof of equivariance for the group convolution of \citet{CohenW16} (see Supplementary material) no longer holds. We cannot prove equivariance of this new expression without an extra property. This property is unitarity of $\mc{L}_g$ with respect to the inner product; that is
\begin{align}
    \sum_{x\in X} \mc{L}_g[f](x) \mc{L}_g[\psi](x) = \sum_{x\in X} f(x) \psi(x).
\end{align}
This leads to the proof of equivariance:
\begin{align}
     [\mc{L}_t[f] \star_G \psi](g) &= \sum_{x \in G} \mc{L}_t[f](x) \mc{L}_g[\psi](x) \\
     &= \sum_{x \in G} f(x) \mc{L}_{t^{-1}g}[\psi](x) \\
     &= [f \star_G \psi](t^{-1}g) \\ 
     &= \mc{L}_t[f \star_G \psi](g) .
\end{align}
From the first to the second lines, we have used the unitarity property; from the second to third lines we have used the definition of the group convolution; and from the third to fourth lines we have used the fact that the group action on functions on the group is defined as $\mc{L}_t[f](g) = f(t^{-1}g)$. We call this version of the group convolution the \emph{unitary group convolution}, to differentiate it from the regular group convolution of Equation \ref{eq:group-convolution-old}.

Unitarity imposes a very intuitive constraint on the form of the group convolution. This constraint is that the response value at $g$ matters only on the relative transformation between $f$ and $\psi$, and is independent of their absolute transformations. Linking back to the group convolution of \citet{CohenW16}, the inner product is automatically unitary if we choose the group action to be of the form
\begin{align}
    \mc{L}_g[f](x) = f(\mc{L}_g^{-1}[x]),
\end{align}
so this is a strict generalization of that work. By demanding unitarity, we see that we cannot use any interpolation method to impose filter transformations. Neither the bilinear interpolation of \citet{BekkersLVEPD18} nor the Fourier-based interpolation of \citet{WeilerHS18} satisfy unitarity.

\subsection{Representing Filter Spaces}
We want to build a convolution without using a proper group, but an arbitrary discretization of a continuous group, which we denote as $\tilde{G}$. Thus we wish to approximate the equivariance properties of the unitary group convolution. This boils down to us having to find an approximate action $\mc{L}$. 

We choose to learn this action based on examples of images and their rotated versions. Furthermore, since we require a technique to rotate all filters independent of their particular values, what we are actually searching for is a technique to rotate an entire filter space. We choose to model filters $\psi$ as living in a linear vector space spanned by a basis $\{e^i(x)\}_{i=1}^N$, where $x \in \tilde{G}$. Then each filter can be formed as a linear combination of the bases:
\begin{align}
    \psi(x) = \sum_{i=1}^N \hat{\psi}_i e^i(x)
\end{align}
where $\{\hat{\psi}_i\}_{i=1}^N$ are the filter coefficients. As a shorthand, we will stack the basis into a vector $e = [e^1(x), ..., e^N(x)]^\top$ and the coefficients into a vector $\hat{\psi} = [\hat{\psi}_1, ..., \hat{\psi}_N]^\top$, so that we can write $\psi(x) = \hat{\psi}^\top e(x)$. Now if we apply an action to the filter, we have
\begin{align}
    \mc{L}_g[\psi](x) &= \mc{L}_g \left [ \hat{\psi}^\top e \right ](x) =  \hat{\psi}^\top \mc{L}_g [e](x)
\end{align}
where in the second equality we have imposed that the action be linear with respect to the input function. In fact linearity is not an extra condition, since unitary operators are always linear. So to transform a filter, we have to find the filter coefficients $\hat{\psi}$ in the original basis $e(x)$ and change the basis for its transformed version $\mc{L}_g[e](x)$. If we can achieve this, then the filter coefficients $\hat{\psi}$ are invariant to the transformation. This is a generalized form of weight-tying.

The main problem now is to find a transformed basis $\mc{L}_g[e](x)$. Instead of learning how to transform the basis, we choose to learn a separate basis for each transformation $g$. We denote this learned basis as $e_g(x)$, so
\begin{align}
    \mc{L}_g[\hat{\psi}^\top e](x) = \hat{\psi}^\top e_g(x)
\end{align}
for all $\hat{\psi}$ and $g$. Algorithmically this basis is used at training time as shown in Algorithm \ref{alg:post-training}. Looking at Equation \ref{eq:intermediate-conv}, we see that the roto-translation convolution contains two parts. One is a spatial rotation, the other is an axis roll. We know exactly how to perform an axis roll, but not how to perform the rotation. We opt to learn the rotation aspect only and hard-code the roll ourselves.

\begin{algorithm}[t]
\SetAlgoLined
    $\text{Load basis: }e$ \\
    $\text{Initialize coefficients: } \hat{\psi}$ \\
\For{minibatch in dataset}{
    $\psi \gets \hat{\psi}^\top e$ \\
    Forward prop and compute loss \\
    $\hat{\psi} \gets \text{backprop update}(\frac{\partial L}{\partial \hat{\psi}})$
}
\caption{Task-specific training using pre-trained basis}
\label{alg:post-training}
\end{algorithm}

\begin{algorithm}[t]
\SetAlgoLined
    $\text{Initialize basis: } e$ \\
    $\text{Set sample } angles = [0, \frac{\pi}{8}, \frac{2\pi}{8}, ... \frac{7\pi}{8}]$ \\
 \For{minibatch in dataset}{
  $\text{Draw } S, R \sim \text{Uniform}(angles)$ \\
  $\text{Compute } L_{\text{equiv}}, L_{\text{rec}}, L_{\text{orth}} \\
  L \gets L_{\text{equiv}} + L_{\text{rec}} + L_{\text{orth}} \\
  e \gets \text{backprop update}(\frac{\partial L}{\partial e})$
 }
 Save basis $e$
\caption{The basis is pretrained offline}
\label{alg:pre-training}
\end{algorithm}

\subsection{Learning To Rotate}
Recall that the roto-translation convolution at the input and intermediate layers can be written
\begin{align}
    [f \star_{G} \psi](R,z) &= [f \star_{\mbb{Z}^d} \mc{R}_R[\psi]](z) \\
    [f \star_{G} \psi](R,z) &= \sum_{S} [f_S \star_{\mbb{Z}^d} \mc{R}_R[\psi_{R^{-1}S}]](z).
\end{align}
From these expressions, we see that the convolution is composed of translations, rotations and rolls. We know how to perform translations and rolls exactly, but not rotations. Thus we only need to learn how to rotate our bases, which we do by minimizing a sum of three losses (explained below),
\begin{align}
    L = L_{\text{equiv}} + L_{\text{orth}} + L_{\text{rec}}.
\end{align}
Algorithmically this process is shown in Algorithm \ref{alg:pre-training}.

\paragraph{Orthogonality loss}
To encourage the learned basis to span the space of filters, we add an orthogonality loss
\begin{align}
    L_{\text{orth}} = \sum_{R}\left \| e_R {e_R}^{\top} - I \right \|_1
\end{align}
for each $R$-rotated version of the basis, where $I$ is an identity matrix of size $N\times N$, and $N$ is number of basis rotations.

\paragraph{Equivariance loss}
We make use of the following fact
\begin{align}
    \mc{R}_S[f] \star_{\mbb{Z}^d} \mc{R}_R[\psi] = \mc{R}_S[f \star_{\mbb{Z}^d} \mc{R}_{S^{-1}R}[\psi]] \label{eq:roto-equivariance}
\end{align}
which is proven in the \emph{Supplementary material}. This relationship shows that translationally convolving an $S$-rotated image $\mc{R}_S[f]$ with an $R$-rotated filter is the same as convolving the unrotated image $f$ with an $S^{-1}R$-rotated filter and then rotating the responses by $S$. The usefulness of Equation \ref{eq:roto-equivariance} is that it connects filters at rotation $R$ with filters at rotation $S^{-1}R$. Thus we can use it to learn rotated versions of filters. Since this expression is correct for any $\psi$, it must also be correct for any basis element $e^i$. To learn a base element and its rotated versions we minimize an $L1$-loss
{
\medmuskip=1mu
\thinmuskip=1mu
\thickmuskip=1mu
\begin{align}
	L_\text{equiv} &= \sum_{S,R,i} L_{\text{equiv}}^{S,R,i} \\
    L_{\text{equiv}}^{S,R,i} &= \sum_{x \in \mbb{Z}^d} \left \| [\mc{R}_S[f] \star_{\mbb{Z}^d} e^i_R] - \mc{R}_S[f \star_{\mbb{Z}^d} e^i_{S^{-1}R} ] \right \|_{1} \label{basis_equiv_loss}
\end{align}
}
for all rotations $S$ and $R$, all basis elements $\{e^i\}_{i=1}^N$, and all $f$ in our training set. $\|\cdot\|_1$ is the $L1$-norm. In practice, we sample angles $S$ and $R$ at random from the set of rotations (in our case $S,R \in \{\frac{\pi n}{4} | n \in \{0,...,7\} \}$.

\paragraph{Reconstruction loss}
Finally, we find that an extra reconstruction loss helps in learning a basis. If we translationally convolve a signal $f$ with a base element $e^i$, we reconstruct the original signal by translationally convolving the response with the transposed filter (flipped in the $x$ and $y$ axes) \cite{DumoulinV2016}, which we denote as $\bar{e}^i$. The reconstruction does not work for any $e^i$, but only for a subset of filters, known as \emph{self-adjoint}. These are in fact unitary. The transpose convolution is denoted as: $\star_{\mbb{Z}^d}\bar{e}$, where we only change the notation for the basis, because when the convolution is viewed as a matrix multiplication, the transpose convolution operation, given the same filters, is equivalent to transposing the matrix of the filters \cite{DumoulinV2016}. If the reconstruction task works then $e^i$ convolved with its transpose $\bar{e}^i$ should result in a Dirac delta, so we have that $e^i \star_{\mbb{Z}^d} \bar{e}^i = \delta(0)$. This allows us to find the following expression
\begin{align}
    \mc{R}_S[f] &= \mc{R}_S[f] \star_{\mbb{Z}^d} [e_R^i \star_{\mbb{Z}^d} \bar{e}_R^i] \\
    &= \mc{R}_S[f \star_{\mbb{Z}^d} e^i_{RS^{-1}}] \star_{\mbb{Z}^d} \bar{e}_R^i
\end{align}
For the reconstruction loss, we found a sum over the different basis elements leads to stabler optimization:
\begin{align}
    L_{\text{rec}} &= \sum_{x \in \mbb{Z}^d} \left \| \mc{R}_S[f] (x) - \sum_i[\mc{R}_S[f \star_{\mbb{Z}^d} e^i_{RS^{-1}}] \star_{\mbb{Z}^d} \bar{e}_R^i] (x) \right \|_1 \label{basis_rec_loss}
\end{align}

\subsection{Implementation details}
\paragraph{Imposing Extra Constraints}
We do not need to learn how to rotate at every angle, since we can rotate by $90^\circ$ perfectly due to the square symmetry of the grid. We thus only learn to rotate in the range $[0^\circ, 90^\circ)$ and populate all other filters, by exactly rotating our learned filters afterwards.

\paragraph{Convolving using a basis}
Once we have learned a basis, we implement the unitary group-like convolution as
\begin{align}
    [f \star_{G} \psi](R,z) &= [f \star_{\mbb{Z}^d} \hat{\psi}^\top e_R](z) \\
    [f \star_{G} \psi](R,z) &= \sum_{S} [f_S \star_{\mbb{Z}^d} \hat{\psi}_{R^{-1}S}^\top e_{R}](z).
\end{align}

\paragraph{Rotating images}
To train the equivariance loss and the reconstruction loss, we still have to perform a spatial rotation of the feature maps. We do this using a Gaussian interpolator with width $\sigma=0.5$ and a kernel of size $3$. Due to the fact that, when rotating an image, we have to interpolate values outside of the image grid, when we compute either the equivariance or the reconstruction, we crop $\frac{1}{4}$ of the feature maps on all sides. 

\section{Experiments and Results}
\label{sec:experiments}
We present our results on some basic benchmarks. We demonstrate competitive classification performance on the CIFAR-10 image recognition dataset, and compare the equivariance properties of activations from different models. 

\subsection{CIFAR-10}
We run a simple test on the CIFAR-10 classification benchmark \cite{Krizhevsky09} to check that our network is able to produce comparable results to standard baselines. We compare against a standard CNN (\textsc{Conv}), a group CNN with a random basis (\textsc{Random}), a group CNN with a Fourier-basis as per \citet{WeilerHS18} (\textsc{Weiler}), a group CNN with a Gaussian-interpolated basis (\textsc{Gaussian}), and a group CNN with a bilinear-interpolated basis (\textsc{Bilinear}). For the last two interpolated basis methods, we also need a `zero-orientation' basis to rotate, so we use one of our learned bases. We also compare three versions of our learned basis: a basis learned at all angles (\textsc{Full}), a basis learned in the range $0^\circ - 90^\circ$ and manually rotated to the other angles (\textsc{Partial}), and an overcomplete basis (\textsc{Overcomplete}) learned at all angles with 3 times as many base elements.

Our model, detailed in the Appendix, is an All-CNN-C-like \citep{SpringenbergDBR14} architecture. For all our experiments we used $|G|=8$, meaning we have 8 orientations of the basis, at multiples of 45 degrees. We used the AMSGrad variant of Adam \citep{ReddiKK18} as the optimizer and we trained for 100 epochs at learning rate $10^{-3}$ and weight decay $10^{-6}$. We use a minibatch size of $100$. For data augmentation, we use random flips, color normalization and random translations of at most 4 pixels. 

\paragraph{Accuracy: CIFAR-10}
The accuracy results can be seen in Table \ref{tab:cifar-results}. When using no augmentation, we can see that interestingly, the random baseline performs very well. Among interpolation methods, ours performs best. Moreover, the difference between translation and roto-translational model increases as we add rotation augmentation to the dataset. The key message we learn from this is that our learned filter basis is competitive with other baseline methods. We can see, by comparison, that there is an overall decrease in performance from the column with no augmentation to the column with full data augmentation. We hypothesize that CIFAR-10 contains an orientation bias (\eg the sky is always up). As a result, the non roto-equivariant models need extra capacity to learn object classifications at all angles, when the roto-equivariant models have this already built in.

\begin{table}[t]
    \centering
    \caption{Results on the CIFAR-10 benchmark, with no augmentation left and with full rotation augmentation (right). The methods are split into no roto-translation equivariance, handcrafted equivariance, and learned equivariance (ours).}
    \label{tab:cifar-results}
    \begin{tabular}{l|c | c}
        \textsc{Method} & \textsc{Test acc} & \textsc{Test acc (aug.)} \\
        \hline \hline
        All-CNN-C-like      & 92.23             & 84.14   \\
        Random              & 90.79             & 80.86  \\
        \hline
        \citet{WeilerHS18}  & 90.2              & 89.48  \\
        Bilinear            & 90.63             & 89.41  \\
        Gaussian            & 90.72             & 89.45  \\
        \hline
        Full (Ours)      & \textbf{92.75}    & 89.84  \\
        Partial (Ours)      & 90.93             & 89.60  \\
        Overcomplete (Ours) & 91.16             & \textbf{89.93}  \\
    \end{tabular}
\end{table}
\subsection{Testing equivariance}
\begin{figure}[t]
    \centering
    \includegraphics[width=\columnwidth]{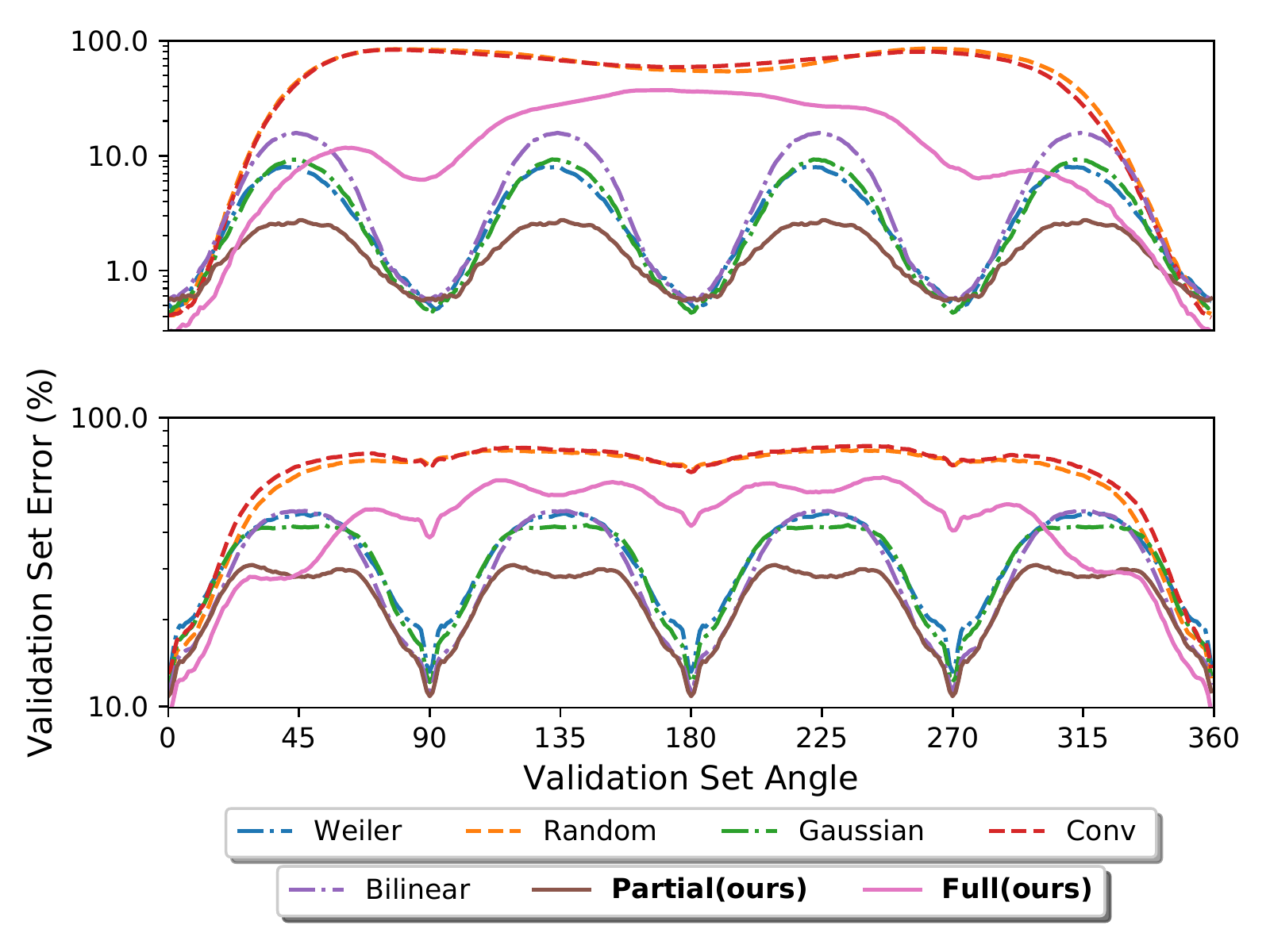}
    \caption{Validation error of a set of models on MNIST (top) and CIFAR-10 (bottom) when we rotate the validation set at different angles. We see similar behaviors in both datasets. Notably, the \textsc{Partial} method performs best.}
    \label{fig:equivariance-plot}
\end{figure}

\begin{figure}[t]
    \centering
    \includegraphics[width=\columnwidth]{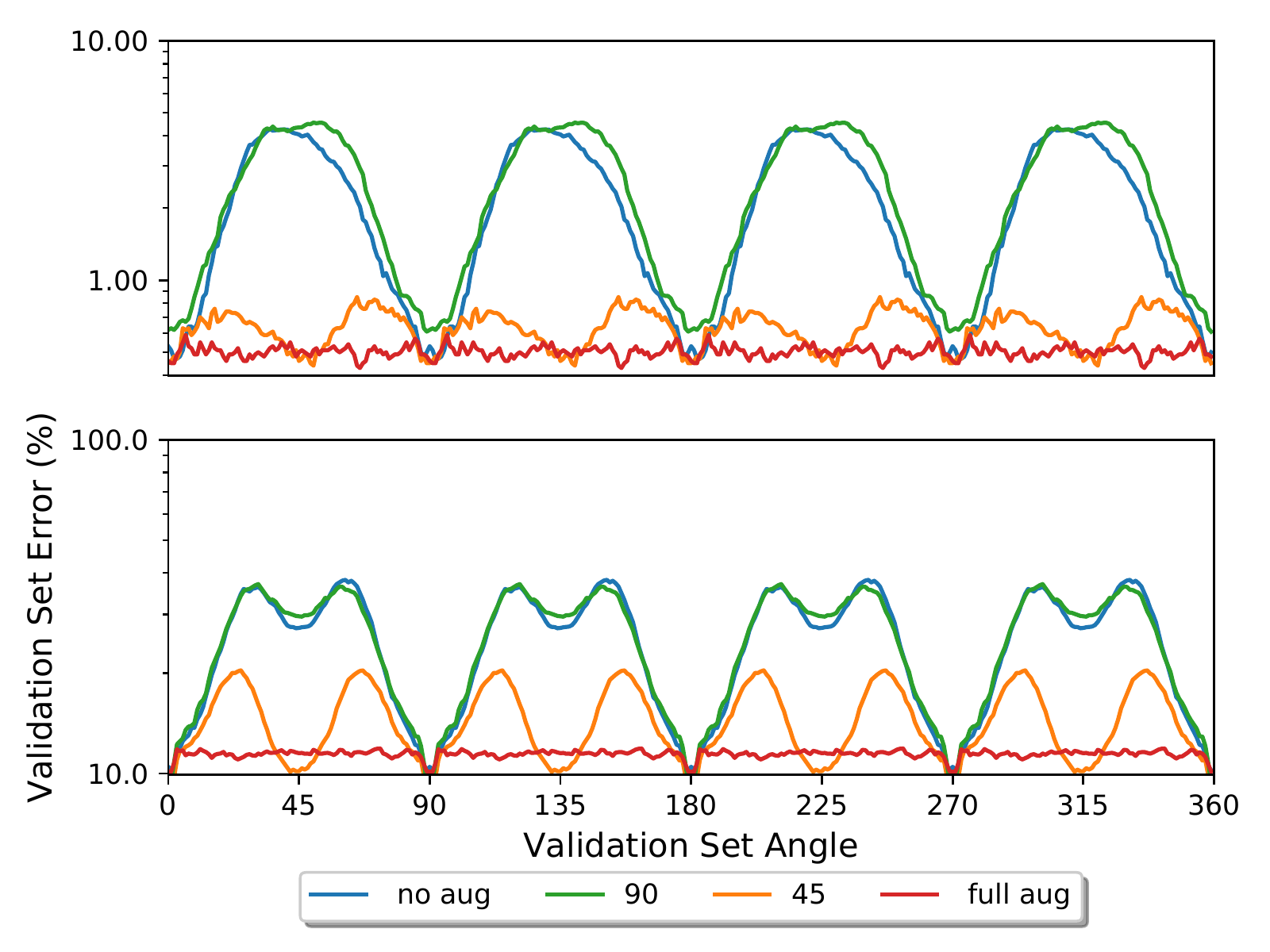}
    \caption{Validation error of the \textsc{Partial} model on MNIST (top) and CIFAR-10 (bottom) when we rotate the validation set at different angles for various degrees of rotational data augmentation during training.}
    \label{fig:augmentation-plot}
\end{figure}

\paragraph{Test error fluctuations}
One test of equivariance is to monitor how test error fluctuates as a function of input rotation. For this we followed a similar procedure to \citet{WeilerHS18} and trained all models on the MNIST and CIFAR-10 classification benchmarks. We also trained the \textsc{Partial} model using various amounts of data augmentation: rotations at multiples of $90^\circ, 45^\circ$ or rotations at all angles in $[0^\circ,360^\circ)$, which we call `full augmentation'. We plot the test error against input image rotation of the test set in Figures \ref{fig:equivariance-plot} and \ref{fig:augmentation-plot}. Our models are shown in solid lines (pink/brown). The dashed lines show translational models and the dotted-and-dashed lines show competing group convolutional models using handcrafted interpolation methods.

\begin{figure}[h]
    \centering
    \includegraphics[width=\columnwidth]{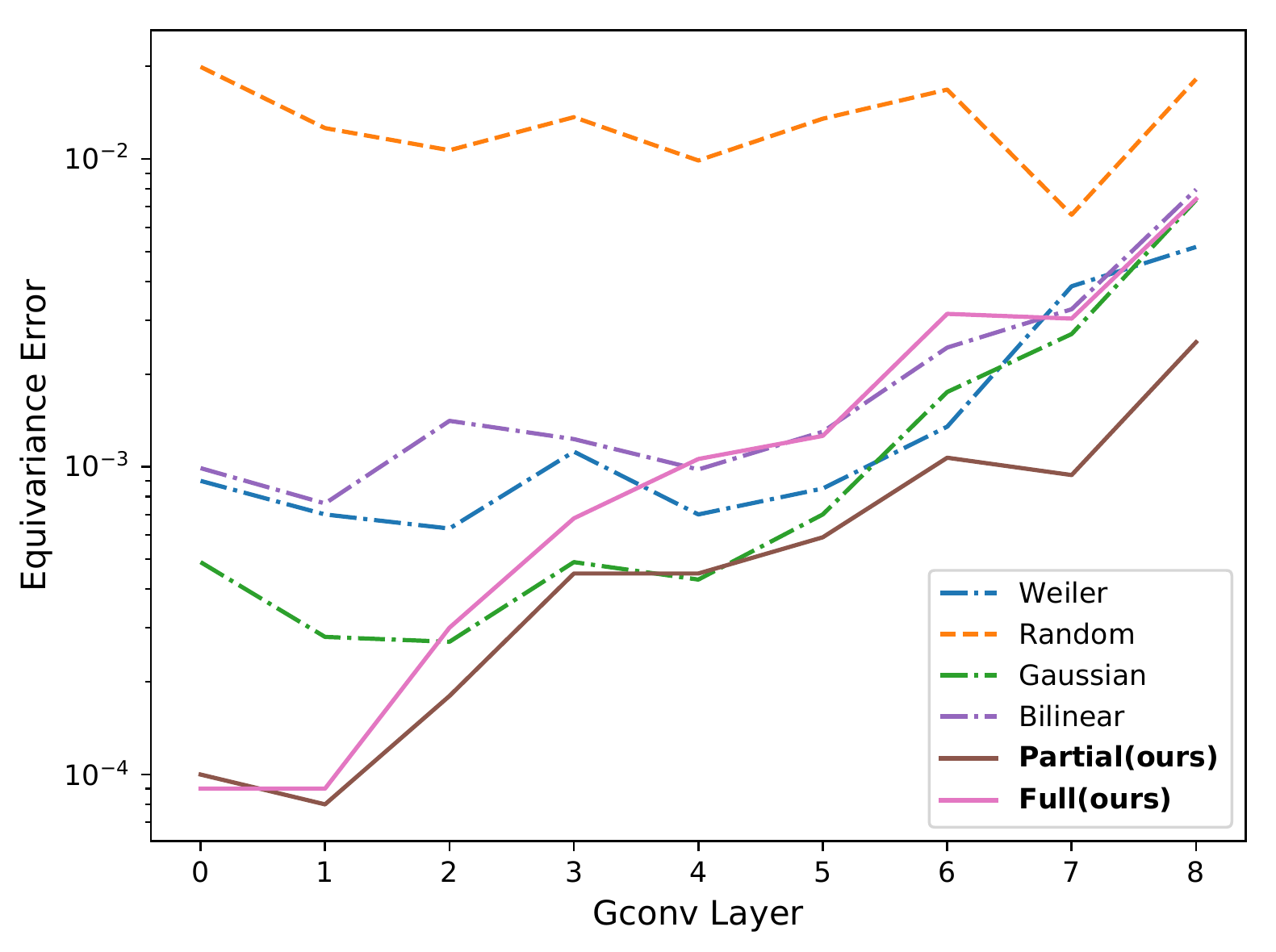}
    \caption{Per layer equivariance loss of the models in the baseline. We see that as the network depth increases all of the equivariance losses also increase. The handcrafted interpolation methods have higher error than the learned ones and this is most pronounced at the input to the network. Among the learned methods, the \textsc{Partial} one performs per than the \textsc{Full} method. We suspect that small asymmetries in the fully learned basis are exploited at training time to overfit to an orientation bias in the data.}
    \label{fig:per-layer-eq}
\end{figure}

In Figures  \ref{fig:equivariance-plot} we see that our \textsc{Partial} method performs best over all runs. At its worst it has a $\sim 2\%$ error increase over its best error on MNIST and $20\%$ on CIFAR-10. By comparison, the other interpolation methods suffer from a $7-10\%$ error increase on MNIST and $30-40\%$ error increase on CIFAR-10. Interestingly, the fully learned basis does not perform well. We posit that small asymmetries in the fully learned basis may be exploited during training time to overfit on a single input orientation. Further evidence to support this is shown in Figure \ref{fig:per-layer-eq}, where we demonstrate that the fully learned basis has poor equivariance error compared to models with exact equivariance to $90^\circ$ rotations, such as the \textsc{Partial} model. In Figure \ref{fig:augmentation-plot} we see that the more augmentation we add into the training set, the more equivariant our model becomes, while also increasing in performance at intermediate angles, between multiples of $45^\circ$. The graph also shows that, for MNIST, the accuracy on the unrotated validation set increases with data augmentation, while on CIFAR-10, the highest accuracy does not change. An exception to this is when adding only $90^\circ$ augmentation, which results in no improvement since \textsc{Partial} is already equivariant to $90^\circ$ rotations.

\paragraph{Activation robustness}
Another way to test for equivariance is to measure fluctuations in the activations under rotation of the input. For this, we took trained networks and fed in pairs of test images. Each pair of test images $(\mc{R}_R[f], \mc{R}_S[f])$ consists of the same image but at different angles defined by $R$ and $S$. To measure the equivariance error, we then have to `rectify' the activations by transforming one of them, say the second image's activations $a_S$, by $\mc{R}_{RS^{-1}}$. We use is the squared normalized $L_2$-error metric
\begin{align}
    L(a_R, a_S)^2 = \sum_k \sum_{x\in \mbb{Z}^2} \frac{\left \| a_R^k(x) - \mc{R}_{RS^{-1}}[a_S^k](x) \right \|_2^2}{\| a_R^k(x) \|_2  \| \mc{R}_{RS^{-1}}[a_S^k](x)\|_2},
\end{align}
where $k$ is the channel number. We set $R = 0$ and average this loss over $N$ images and all angles $S$ in the set of rotations, where $N_R$ is number of rotations, so
\begin{align}
    L_{\text{equivariance}} = \frac{1}{N N_R}\sum_{a}\sum_{R} L(a_0, a_R)^2.
\end{align}
We show the errors per layer in Figure \ref{fig:per-layer-eq}. We see that the \textsc{Partial} method performs best. We also see that the \textsc{Full} method is similar to the \textsc{Partial} method in terms of equivariance error for the first two layers. We suspect small asymmetries in the \textsc{Full} basis make the errors build up over depth, so that at the final layer the equivariance loss reaches the same values as the \textsc{Bilinear} or \textsc{Gaussian} methods. Our experiments demonstrate that with a finite number of rotations we can achieve good equivariance properties. 

\section{Related Work}
There are a number of recent works on group convolutions, these are: continuous roto-translation in 2D \cite{WorrallGTB17} and 3D \cite{WeilerGWBC18, KondorLT18, ThomasSKYLKR18} and discrete roto-translations in 2D \cite{CohenW16, WeilerHS18, BekkersLVEPD18, HoogeboomPCW18} and 3D \cite{WorrallGTB17}, continuous rotations on the sphere \cite{EstevesAMD18, CohenGKW18}, in-plane reflections \cite{CohenW16}. All of these works use handcrafted filter transformations. The closest to ours are \citet{BekkersLVEPD18} and \citet{WeilerHS18} who handcraft interpolation methods to deal with \emph{post hoc} discretization of the roto-translation group. \citet{SabourFH17} and \citet{HintonSF18} are the only other known works by the authors, where equivariance is learned as part of the objective function.

\citet{JacobsenGLS16} and \citet{WeilerHS18} both represent filters using smooth bases to rotate filters by discrete angles; whereas \citet{WorrallGTB17}, \citet{WeilerGWBC18}, \citet{KondorT18}, and \citet{ThomasSKYLKR18} use the steerability properties of harmonic bases to achieve continuous equivariance. The original work on steerable bases was performed by \citet{FreemanA91} and later formalized by \citet{TeoH97}. In a different line of work \citet{stanley2019designing} propose the HyperNEAT framework, which is an evolutionary algorithm for learning structured network weights. HyperNEAT has the capacity for learning invariances, but it could possibly achieve this without the need for weight-tying. Nonetheless, there is no explicit mechanism encouraging the framework to learn equi/invariance to symmetries in the input. 

\section{Discussion and Future Works}
We have shown that \emph{post hoc} discretization introduces asymmetries in group CNNs and we have tackled this by learning how to rotate filters. We showed that this give us comparable if not better performance on standard benchmarks and that this greatly improves robustness properties of group CNNs to input transformations. At the moment, the pipeline we have constructed is not an end-to-end learning system. Instead, we have to learn the bases offline, and then learn the filter coefficient afterwards. It would be most desirable for us to be able to construct a framework where we could learn both jointly, perhaps in the same vein as capsule networks \cite{SabourFH17}. It would also be interesting to see whether we could learn other kinds of transformation, such as scalings or affine warps, under \emph{post hoc} discretization of non-compact groups, which to date have not be researched in great depth. Ideally, in the future, we should not even have to specify the exact mechanism of transformation but it would be desirable to learn the symmetries automatically. To date, the authors are not aware of any literature, where a CNN learns symmetries in the task from an equivariance standpoint.

\section{Conclusion}
In this work we have made two main contributions. The first was a generalization of the discrete group convolution of \citet{CohenW16} to allow for convolutions over groups, where the transformation operator applied to the kernel is allowed to be more complicated than a simple permutation of the pixel locations. This new model allows for pixel interpolation and is thus suited for non-$90^\circ$ filter rotations. For this generalized group convolution to work, we noted that the transformations have to be unitary under the inner product of the convolution. This indicated that prior works where filters are rotated using bilinear/Fourier-based interpolation do not satisfy the condition of equivariance.

Our second contribution was to introduce a method to learn how to rotate filters, to deal with \emph{post hoc} discretization artifect. This method represents filters as linear combinations of basis filters. To rotate a filter, we swap the basis for a learned, rotated version of the basis. As such, this corresponds to a generalized version of weight-tying. We showed that this method produces competitive accuracy on the CIFAR-10 image recognition benchmark against matched architectures. We also showed that this method outperforms all other methods for representation stability.

\newpage
\bibliography{example_paper}

\begin{thebibliography}{27}
\providecommand{\natexlab}[1]{#1}
\providecommand{\url}[1]{\texttt{#1}}
\expandafter\ifx\csname urlstyle\endcsname\relax
  \providecommand{\doi}[1]{doi: #1}\else
  \providecommand{\doi}{doi: \begingroup \urlstyle{rm}\Url}\fi

\bibitem[Bekkers et~al.(2018)Bekkers, Lafarge, Veta, Eppenhof, Pluim, and
  Duits]{BekkersLVEPD18}
Bekkers, E.~J., Lafarge, M.~W., Veta, M., Eppenhof, K. A.~J., Pluim, J. P.~W.,
  and Duits, R.
\newblock Roto-translation covariant convolutional networks for medical image
  analysis.
\newblock In \emph{Medical Image Computing and Computer Assisted Intervention -
  {MICCAI} 2018 - 21st International Conference, Granada, Spain, September
  16-20, 2018, Proceedings, Part {I}}, pp.\  440--448, 2018.
\newblock \doi{10.1007/978-3-030-00928-1\_50}.

\bibitem[Cohen \& Welling(2016{\natexlab{a}})Cohen and Welling]{CohenW16}
Cohen, T. and Welling, M.
\newblock Group equivariant convolutional networks.
\newblock In \emph{Proceedings of the 33nd International Conference on Machine
  Learning, {ICML} 2016, New York City, NY, USA, June 19-24, 2016}, pp.\
  2990--2999, 2016{\natexlab{a}}.

\bibitem[Cohen \& Welling(2016{\natexlab{b}})Cohen and Welling]{CohenW16a}
Cohen, T.~S. and Welling, M.
\newblock Steerable cnns.
\newblock \emph{CoRR}, abs/1612.08498, 2016{\natexlab{b}}.

\bibitem[Cohen et~al.(2018)Cohen, Geiger, K{\"{o}}hler, and
  Welling]{CohenGKW18}
Cohen, T.~S., Geiger, M., K{\"{o}}hler, J., and Welling, M.
\newblock Spherical cnns.
\newblock \emph{CoRR}, abs/1801.10130, 2018.

\bibitem[Cordts et~al.(2016)Cordts, Omran, Ramos, Rehfeld, Enzweiler, Benenson,
  Franke, Roth, and Schiele]{CordtsORREBFRS16}
Cordts, M., Omran, M., Ramos, S., Rehfeld, T., Enzweiler, M., Benenson, R.,
  Franke, U., Roth, S., and Schiele, B.
\newblock The cityscapes dataset for semantic urban scene understanding.
\newblock In \emph{2016 {IEEE} Conference on Computer Vision and Pattern
  Recognition, {CVPR} 2016, Las Vegas, NV, USA, June 27-30, 2016}, pp.\
  3213--3223, 2016.
\newblock \doi{10.1109/CVPR.2016.350}.

\bibitem[{Dumoulin} \& {Visin}(2016){Dumoulin} and {Visin}]{DumoulinV2016}
{Dumoulin}, V. and {Visin}, F.
\newblock {A guide to convolution arithmetic for deep learning}.
\newblock \emph{arXiv e-prints}, art. arXiv:1603.07285, March 2016.

\bibitem[Esteves et~al.(2018)Esteves, Allen{-}Blanchette, Makadia, and
  Daniilidis]{EstevesAMD18}
Esteves, C., Allen{-}Blanchette, C., Makadia, A., and Daniilidis, K.
\newblock Learning {SO(3)} equivariant representations with spherical cnns.
\newblock In \emph{Computer Vision - {ECCV} 2018 - 15th European Conference,
  Munich, Germany, September 8-14, 2018, Proceedings, Part {XIII}}, pp.\
  54--70, 2018.
\newblock \doi{10.1007/978-3-030-01261-8\_4}.

\bibitem[Freeman \& Adelson(1991)Freeman and Adelson]{FreemanA91}
Freeman, W.~T. and Adelson, E.~H.
\newblock The design and use of steerable filters.
\newblock \emph{{IEEE} Trans. Pattern Anal. Mach. Intell.}, 13\penalty0
  (9):\penalty0 891--906, 1991.
\newblock \doi{10.1109/34.93808}.

\bibitem[Hinton et~al.(2018)Hinton, Sabour, and Frosst]{HintonSF18}
Hinton, G., Sabour, S., and Frosst, N.
\newblock Matrix capsules with em routing.
\newblock 2018.
\newblock URL \url{https://openreview.net/pdf?id=HJWLfGWRb}.

\bibitem[Hoogeboom et~al.(2018)Hoogeboom, Peters, Cohen, and
  Welling]{HoogeboomPCW18}
Hoogeboom, E., Peters, J. W.~T., Cohen, T.~S., and Welling, M.
\newblock Hexaconv.
\newblock \emph{CoRR}, abs/1803.02108, 2018.

\bibitem[Jacobsen et~al.(2016)Jacobsen, van Gemert, Lou, and
  Smeulders]{JacobsenGLS16}
Jacobsen, J., van Gemert, J.~C., Lou, Z., and Smeulders, A. W.~M.
\newblock Structured receptive fields in cnns.
\newblock In \emph{2016 {IEEE} Conference on Computer Vision and Pattern
  Recognition, {CVPR} 2016, Las Vegas, NV, USA, June 27-30, 2016}, pp.\
  2610--2619, 2016.
\newblock \doi{10.1109/CVPR.2016.286}.

\bibitem[Kondor \& Trivedi(2018)Kondor and Trivedi]{KondorT18}
Kondor, R. and Trivedi, S.
\newblock On the generalization of equivariance and convolution in neural
  networks to the action of compact groups.
\newblock In \emph{Proceedings of the 35th International Conference on Machine
  Learning, {ICML} 2018, Stockholmsm{\"{a}}ssan, Stockholm, Sweden, July 10-15,
  2018}, pp.\  2752--2760, 2018.

\bibitem[Kondor et~al.(2018)Kondor, Lin, and Trivedi]{KondorLT18}
Kondor, R., Lin, Z., and Trivedi, S.
\newblock Clebsch-gordan nets: a fully fourier space spherical convolutional
  neural network.
\newblock In \emph{Advances in Neural Information Processing Systems 31: Annual
  Conference on Neural Information Processing Systems 2018, NeurIPS 2018, 3-8
  December 2018, Montr{\'{e}}al, Canada.}, pp.\  10138--10147, 2018.

\bibitem[Krizhevsky(2009)]{Krizhevsky09}
Krizhevsky, A.
\newblock Learning multiple layers of features from tiny images.
\newblock Technical report, 2009.

\bibitem[Reddi et~al.(2018)Reddi, Kale, and Kumar]{ReddiKK18}
Reddi, S.~J., Kale, S., and Kumar, S.
\newblock On the convergence of adam and beyond.
\newblock In \emph{International Conference on Learning Representations}, 2018.

\bibitem[Russakovsky et~al.(2015)Russakovsky, Deng, Su, Krause, Satheesh, Ma,
  Huang, Karpathy, Khosla, Bernstein, Berg, and Li]{RussakovskyDSKS15}
Russakovsky, O., Deng, J., Su, H., Krause, J., Satheesh, S., Ma, S., Huang, Z.,
  Karpathy, A., Khosla, A., Bernstein, M.~S., Berg, A.~C., and Li, F.
\newblock Imagenet large scale visual recognition challenge.
\newblock \emph{International Journal of Computer Vision}, 115\penalty0
  (3):\penalty0 211--252, 2015.
\newblock \doi{10.1007/s11263-015-0816-y}.

\bibitem[Sabour et~al.(2017)Sabour, Frosst, and Hinton]{SabourFH17}
Sabour, S., Frosst, N., and Hinton, G.~E.
\newblock Dynamic routing between capsules.
\newblock In \emph{Advances in Neural Information Processing Systems 30: Annual
  Conference on Neural Information Processing Systems 2017, 4-9 December 2017,
  Long Beach, CA, {USA}}, pp.\  3859--3869, 2017.

\bibitem[Sokolic et~al.(2017)Sokolic, Giryes, Sapiro, and
  Rodrigues]{SokolicGSR17}
Sokolic, J., Giryes, R., Sapiro, G., and Rodrigues, M. R.~D.
\newblock Generalization error of invariant classifiers.
\newblock In \emph{Proceedings of the 20th International Conference on
  Artificial Intelligence and Statistics, {AISTATS} 2017, 20-22 April 2017,
  Fort Lauderdale, FL, {USA}}, pp.\  1094--1103, 2017.

\bibitem[Springenberg et~al.(2014)Springenberg, Dosovitskiy, Brox, and
  Riedmiller]{SpringenbergDBR14}
Springenberg, J.~T., Dosovitskiy, A., Brox, T., and Riedmiller, M.~A.
\newblock Striving for simplicity: The all convolutional net.
\newblock \emph{CoRR}, abs/1412.6806, 2014.

\bibitem[Stanley et~al.(2019)Stanley, Clune, Lehman, and
  Miikkulainen]{stanley2019designing}
Stanley, K.~O., Clune, J., Lehman, J., and Miikkulainen, R.
\newblock Designing neural networks through neuroevolution.
\newblock \emph{Nature Machine Intelligence}, 1\penalty0 (1):\penalty0 24--35,
  2019.

\bibitem[Szeliski(2011)]{Szeliski11}
Szeliski, R.
\newblock \emph{Computer Vision - Algorithms and Applications}.
\newblock Texts in Computer Science. Springer, 2011.
\newblock ISBN 978-1-84882-934-3.
\newblock \doi{10.1007/978-1-84882-935-0}.

\bibitem[Teo \& Hel{-}Or(1997)Teo and Hel{-}Or]{TeoH97}
Teo, P.~C. and Hel{-}Or, Y.
\newblock A computational approach to steerable functions.
\newblock In \emph{1997 Conference on Computer Vision and Pattern Recognition
  {(CVPR} '97), June 17-19, 1997, San Juan, Puerto Rico}, pp.\  313--318, 1997.
\newblock \doi{10.1109/CVPR.1997.609341}.

\bibitem[Thomas et~al.(2018)Thomas, Smidt, Kearnes, Yang, Li, Kohlhoff, and
  Riley]{ThomasSKYLKR18}
Thomas, N., Smidt, T., Kearnes, S.~M., Yang, L., Li, L., Kohlhoff, K., and
  Riley, P.
\newblock Tensor field networks: Rotation- and translation-equivariant neural
  networks for 3d point clouds.
\newblock \emph{CoRR}, abs/1802.08219, 2018.

\bibitem[Weiler et~al.(2018{\natexlab{a}})Weiler, Geiger, Welling, Boomsma, and
  Cohen]{WeilerGWBC18}
Weiler, M., Geiger, M., Welling, M., Boomsma, W., and Cohen, T.
\newblock 3d steerable cnns: Learning rotationally equivariant features in
  volumetric data.
\newblock In \emph{Advances in Neural Information Processing Systems 31: Annual
  Conference on Neural Information Processing Systems 2018, NeurIPS 2018, 3-8
  December 2018, Montr{\'{e}}al, Canada.}, pp.\  10402--10413,
  2018{\natexlab{a}}.

\bibitem[Weiler et~al.(2018{\natexlab{b}})Weiler, Hamprecht, and
  Storath]{WeilerHS18}
Weiler, M., Hamprecht, F.~A., and Storath, M.
\newblock Learning steerable filters for rotation equivariant cnns.
\newblock In \emph{2018 {IEEE} Conference on Computer Vision and Pattern
  Recognition, {CVPR} 2018, Salt Lake City, UT, USA, June 18-22, 2018}, pp.\
  849--858, 2018{\natexlab{b}}.

\bibitem[Worrall \& Brostow(2018)Worrall and Brostow]{WorrallB18}
Worrall, D.~E. and Brostow, G.~J.
\newblock Cubenet: Equivariance to 3d rotation and translation.
\newblock In \emph{Computer Vision - {ECCV} 2018 - 15th European Conference,
  Munich, Germany, September 8-14, 2018, Proceedings, Part {V}}, pp.\
  585--602, 2018.
\newblock \doi{10.1007/978-3-030-01228-1\_35}.

\bibitem[Worrall et~al.(2017)Worrall, Garbin, Turmukhambetov, and
  Brostow]{WorrallGTB17}
Worrall, D.~E., Garbin, S.~J., Turmukhambetov, D., and Brostow, G.~J.
\newblock Harmonic networks: Deep translation and rotation equivariance.
\newblock In \emph{2017 {IEEE} Conference on Computer Vision and Pattern
  Recognition, {CVPR} 2017, Honolulu, HI, USA, July 21-26, 2017}, pp.\
  7168--7177, 2017.
\newblock \doi{10.1109/CVPR.2017.758}.

\end{thebibliography}
\bibliographystyle{icml2019}

\clearpage

\appendix
\title{Appendix}
\maketitle

Here we provide proofs of the equivariance properties of the of the group convolution and unitary group convolution. We also provide our architecture and examples of filters and activations.

\section{Equivariance Of The Group Convolution}
Here we provide a copy of \citet{CohenW16}'s equivariance proof of the discrete group convolution. For a signal $f:X \to \mbb{R}$, filter $\psi:X \to \mbb{R}$, domain $X$, group $G$, and group action $\mc{L}_g$ where $\mc{L}_g[f](x) = f(\mc{L}_{g}^{-1}[x])$, we have
\begin{align}
    [\mc{L}_t[f] \star_G \psi](g) &= \sum_{x \in X} \mc{L}_t[f](x) \psi(\mc{L}_g^{-1}[x]) \\
    &= \sum_{x \in G} f(\mc{L}_{t}^{-1}[x]) \psi(\mc{L}_g^{-1}[x]) \\
    &= \sum_{x' \in G} f(x') \psi(\mc{L}_g^{-1}[\mc{L}_t[x']]) \\
    &= \sum_{x' \in G} f(x') \psi(\mc{L}_{g^{-1}t}[x']]) \\
    &= \sum_{x' \in G} f(x') \psi(\mc{L}_{(t^{-1}g)^{-1}}[x']]) \\
    &= [f \star_G \psi](t^{-1}g) \\
    &= \mc{L}_t[f \star_G \psi](g)
\end{align}
From line 1 to 2 we used the definition $\mc{L}_g[f](x) = f(\mc{L}_{g}^{-1}[x])$; from line 2 to 3 we performed as change of variables $x'=\mc{L}_t^{-1}[x]$ or equally $x = \mc{L}_t[x']$; from line 3 to 4 we applied the composition rule for actions; from line 4 to 5 we used the rule $(ab)^{-1} = b^{-1}a^{-1}$ and in the remaining lines we used the definitions of the group convolution and actions.

\section{The Equivariance Loss}
In the equivariance loss we make use of the following statement
\begin{align}
    \mc{R}_S[f] \star_{\mbb{Z}^d} \mc{R}_R[\psi] = \mc{R}_S[f \star_{\mbb{Z}^d} \mc{R}_{S^{-1}R}[\psi]].
\end{align}
The derivation is as follows. We begin by noting that the roto-translation operator can be written $\mc{L}_{R,z} = \mc{T}_z \mc{R}_R$, where $\mc{T}_z$ is the translation operator and $\mc{R}_R$ is the rotation operator. Then we consider the convolution of an $S$-rotated image $\mc{R}_S[f]$ and filters $\psi$
\begin{align}
    [\mc{R}_S[f] \star_{G}\psi](R,z) &= \sum_{x \in G} \mc{R}_S[f](x) \mc{T}_z[\mc{R}_R[\psi]](x) \\
    &= [\mc{R}_{S}[f](x) \star_{G} \mc{R}_R[\psi]](x)
\end{align}
which constitutes the LHS of the expression. Now for the RHS.
\begin{align}
\mc{R}_S[f] &\star_{\mbb{Z}^d} \mc{R}_R[\psi] \\
    &= [\mc{L}_{S,0}[f] \star_{G} \psi](R,z) \\
    &= \sum_{x \in G} \mc{L}_{S,0}[f](x) \mc{L}_{R,z} [\psi](x) \\
    &= \sum_{x \in G} f(x) \mc{L}_{(S,0)^{-1}}[\mc{L}_{R,z} [\psi]](x) \\
    &= \sum_{x \in G} f(x) \mc{L}_{(S^{-1},0)}[\mc{L}_{R,z} [\psi]](x) \\
    &= \sum_{x \in G} f(x) \mc{L}_{(S^{-1}R,S^{-1}z)}[\psi](x) \\
    &= \sum_{x \in G} f(x) \mc{L}_{(S^{-1}R,S^{-1}z)}[\psi](x) \\
    &= \sum_{x \in G} f(x) \mc{T}_{S^{-1}z} [\mc{R}_{S^{-1}R}[\psi]](x) \\
    &= \sum_{x \in G} f(x) \mc{R}_{S^{-1}R}[\psi](x - S^{-1}z) \\
    &= [f \star_{\mbb{Z}^{d}} \mc{R}_{S^{-1}R}[\psi]](S^{-1}z) \\
    &= \mc{R}_S[f \star_{\mbb{Z}^{d}} \mc{R}_{S^{-1}R}[\psi]](z)
\end{align}
which constitutes the RHS of the expression.

\newpage
\section{Architecture}
\begin{table}[H]
    \centering
    \caption{The architectures of the translational and roto-translational equivariant models. After every convolution we place a batch normalization layer and a ReLU nonlinearity. Across the two models we have fixed the number of channels, such that the number of parameters is roughly the same. `conv$N$' stands for a standard translational convolution of size $N\times N$ and `Gconv$N$' stands for a roto-translational group convolution. Horizontal lines correspond to max pooling of kernel size 2 and stride 2. The global max pool corresponds to a max pool over the spatial dimensions and the orientation dimensions of the activation tensor.}
    \label{tab:model}
    \begin{tabular}{c|c}
        \textsc{Translational} & \textsc{Roto-translational} \\
        \hline
        conv3-96 & Gconv-33 \\
        conv3-96 & Gconv-33 \\
        conv3-96 & Gconv-33 \\
        \hline
        conv3-192 & Gconv-67 \\
        conv3-192 & Gconv-67 \\
        conv3-192 & Gconv-67 \\
        \hline
        conv3-192 & Gconv-67 \\
        conv1-192 & Gconv-67 \\
        conv1-192 & Gconv-67 \\
        global max pool & global max pool \\
        softmax-layer   & softmax-layer
    \end{tabular}
\end{table}

\subsection{Visualization of bases and reconstructions}
\begin{figure}[h]
    \centering
    \includegraphics[width=\columnwidth]{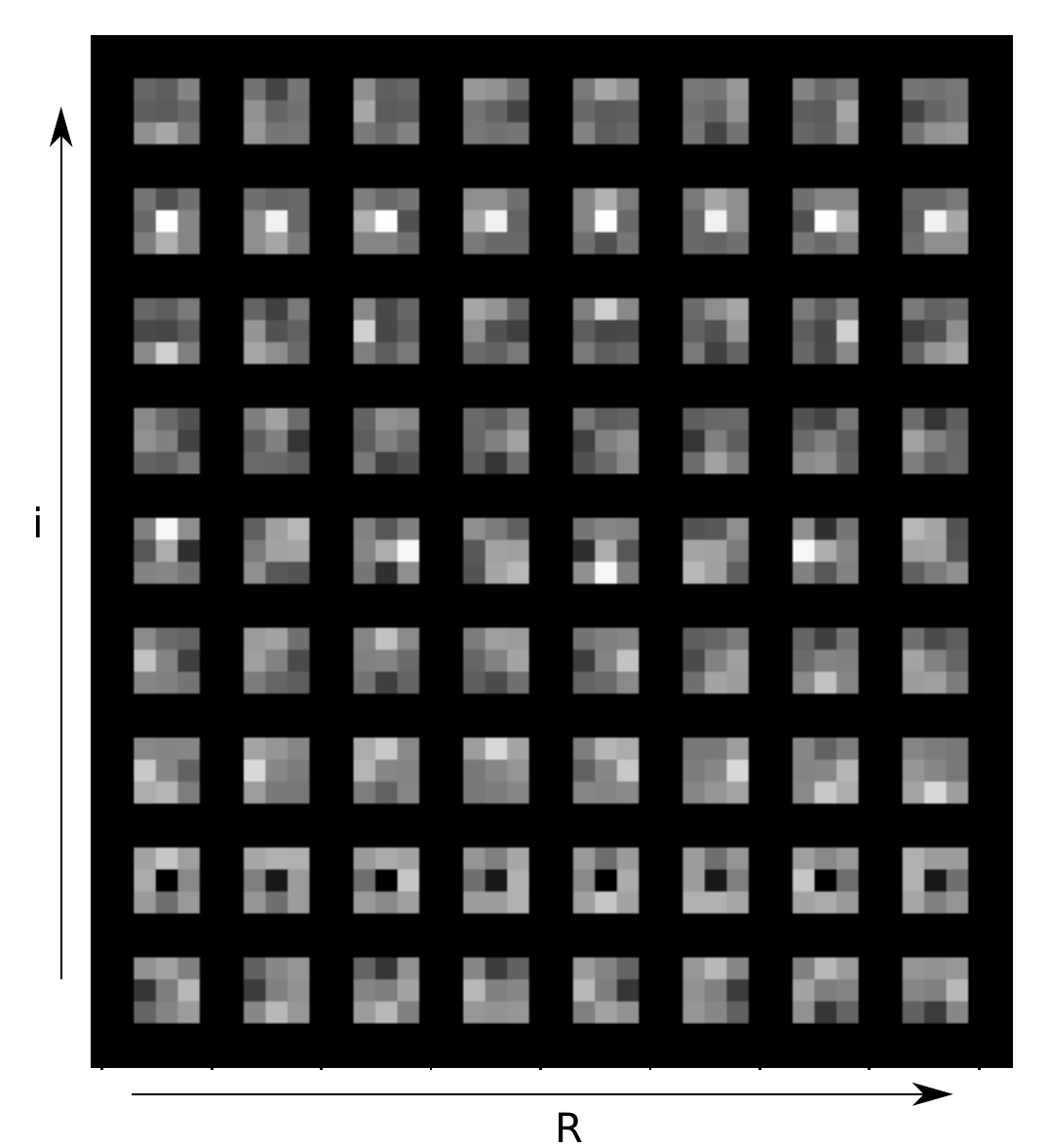}
    \caption{A basis with 9 elements at 8 orientations from an \textsc{Partial} model. $\{e^i_R\}_{i,R}$}
    \label{fig:basis}
\end{figure}
\begin{figure}[h]
    \centering
    \includegraphics[width=\columnwidth]{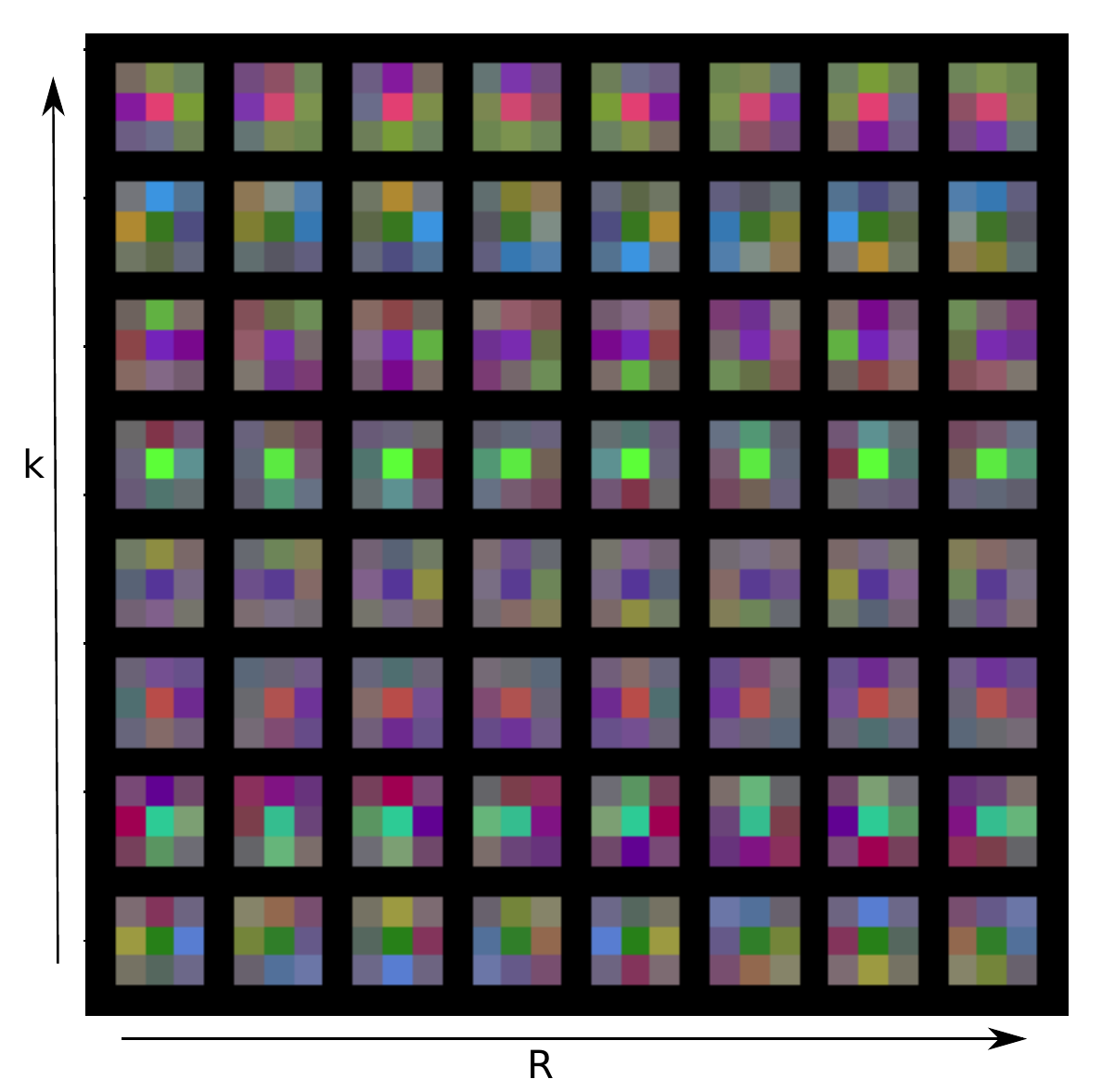}
    \caption{A set of filters from the first layer of an \textsc{Partial} model. $\mathcal{R}_R[\psi_k]$}
    \label{fig:filters_l1}
\end{figure}

\begin{figure}[h]
    \centering
    \includegraphics[width=\columnwidth]{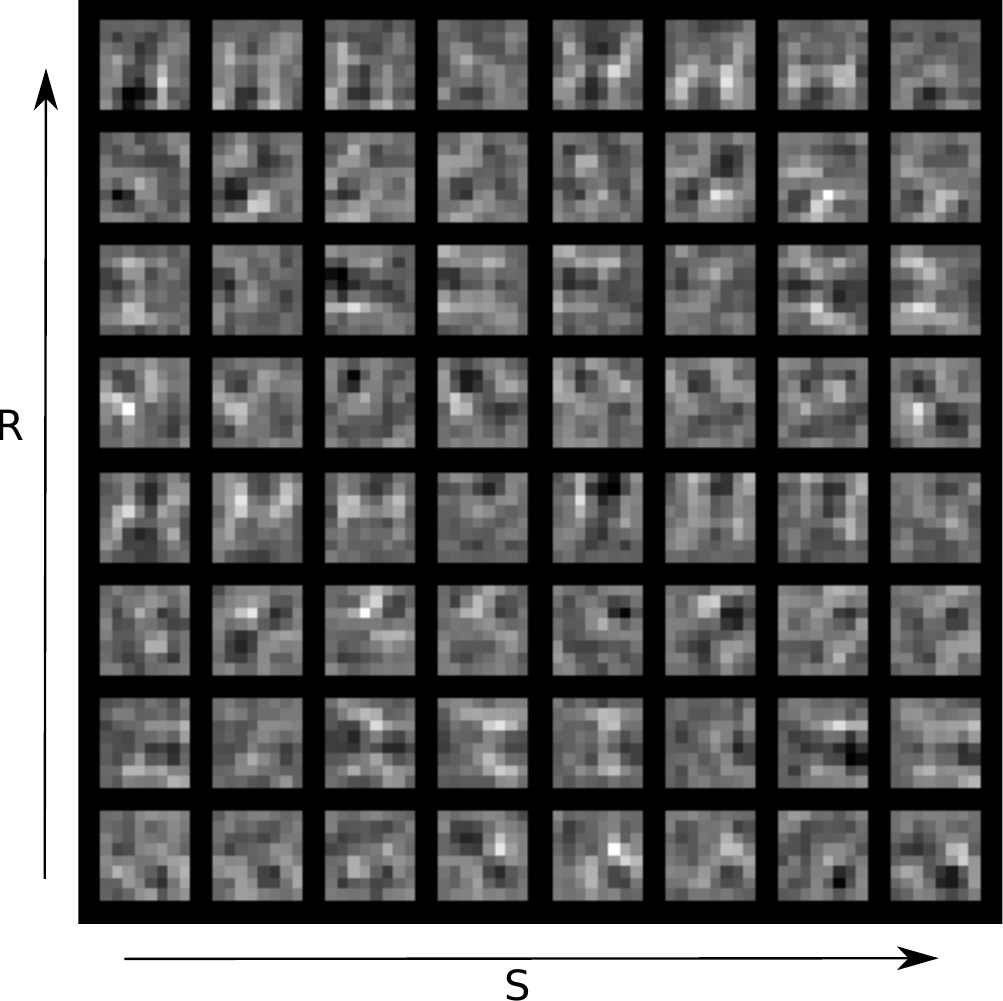}
    \caption{A set activations from an \textsc{Partial} model's layer 6. $\mathcal{R}_R[f]\star_{\mbb{Z}^d} \mathcal{R}_S[\psi]$}
    \label{fig:activations}
\end{figure}
\begin{figure}[h]
    \centering
    \includegraphics[width=\columnwidth]{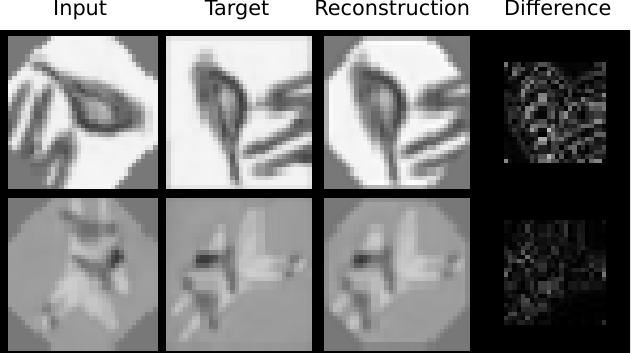}
    \caption{A set of 2 pairs from the reconstruction task, when training the basis. The loss is normalized to the scale of the loss, otherwise it would be too small to distinguish anything.}
    \label{fig:reconstructions-2}
\end{figure}
\begin{figure}[t]
    \centering
    \includegraphics[width=\columnwidth]{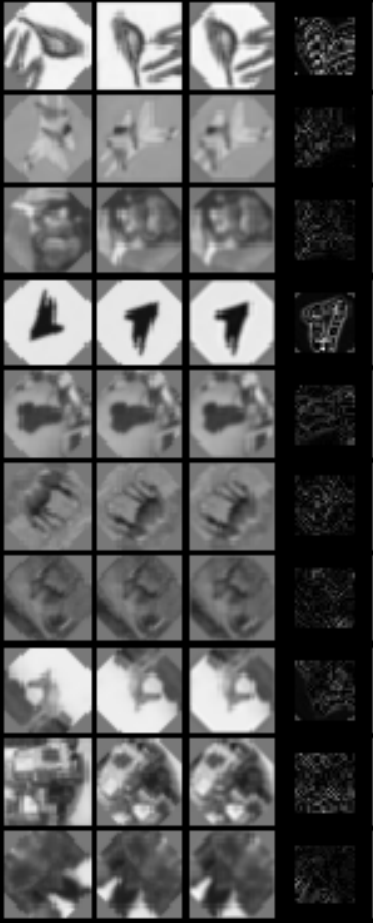}
    \caption{A set of 10 pairs from the reconstruction task, when training the basis. The columns represent in order: the input, the target, the reconstruction, the loss.}
    \label{fig:reconstructions-all}
\end{figure}

\end{document}